\documentclass{article}
\usepackage{subcaption}
\usepackage{spconf,amsmath,amssymb,graphicx,hyperref}

\title{Personal Attribute Leakage in Federated Speech Models}

\name{%
\begin{tabular}{c}
Hamdan Al-Ali\textsuperscript{1}, 
Ali Reza Ghavamipour\textsuperscript{2}, 
Tommaso Caselli\textsuperscript{3} \\
Fatih Turkmen\textsuperscript{3}, 
Zeerak Talat\textsuperscript{4}, 
Hanan Aldarmaki\textsuperscript{1}
\end{tabular}
}

\address{
    $^{1}$Mohamed bin Zayed University of Artificial Intelligence, UAE \\
    $^{2}$Maastricht University, Netherlands 
    ~$^{3}$University of Groningen, Netherlands \\
    $^{4}$University of Edinburgh, UK
}

\begin{document}

\maketitle
\begin{abstract}
Federated learning is a common method for privacy-preserving training of machine learning models. In this paper, we analyze the vulnerability of ASR models to attribute inference attacks in the federated setting. We test a non-parametric white-box attack method under a passive threat model on three ASR models: Wav2Vec2, HuBERT, and Whisper. The attack operates solely on weight differentials without access to raw speech from target speakers. We demonstrate attack feasibility on sensitive demographic and clinical attributes: gender, age, accent, emotion, and dysarthria. Our findings indicate that  attributes that are underrepresented or absent in the pre-training data are more vulnerable to such inference attacks. In particular, information about accents can be reliably inferred from all models. Our findings expose previously undocumented vulnerabilities in federated ASR models and offer insights towards improved security.
\end{abstract}


\section{Introduction}
\label{sec:intro}
Speech technologies are increasingly integrated into daily life, powering applications from virtual assistants to healthcare. These systems rely on machine learning models trained on raw speech to extract acoustic features, enabling robust recognition, synthesis, and understanding across diverse environments and speaker populations. Self-supervised acoustic models—--such as Wav2Vec2~\cite{baevski2020wav2vec} and HuBERT~\cite{hsu2021hubert}--have become central to modern speech processing tasks, including automatic speech recognition (ASR), due to their ability to learn rich representations from unlabeled audio data. Large pre-trained semi-supervised models, like Whisper~\cite{radford2023robust}, are also widely used. These models often need fine-tuning on target speaker data for optimal performance, which raises privacy concerns given the sensitive nature of speech. A single voice sample can inadvertently disclose a wide range of personal attributes—including demographic characteristics (e.g., gender \cite{leung2018voice}, age \cite{mulac1996your}), and accent \cite{clopper2004effects}), clinical conditions (e.g., dysarthria \cite{shih2022dysarthria}), and emotional states. Federated learning (FL) offers privacy-aware training for such data-sensitive domains by keeping raw audio on user devices and sharing only model updates~\cite{mcmahan2017communication}. This approach is well-suited to speech, which is naturally generated on personal devices, and addresses both privacy and scalability concerns. 
However, FL is not immune to threats~\cite{tomashenko2022privacy}. Even without raw inputs, model updates can leak sensitive information~\cite{shokri2017membership, melis2019exploiting}. In the speech domain, prior work has demonstrated speaker re-identification and membership inference against federated ASR models using small samples of target user speech~\cite{ChenZS24, tseng2021membership}. Beyond these membership attacks, the extent of attribute leakage from model weights remains underexplored. In this work, we investigate the risk of inferring personal attributes from weight updates in FL settings, without access to any target user data. Inferring personal attributes from weights alone could enable profiling, surveillance, or discrimination, and may violate legal protections such as GDRP, HIPAA, and anti-discrimination laws in the US and EU. Additionally, the ADA protects people with disabilities from discrimination \cite{MOHAMMED2025100215}. As noted by~\cite{ritter2021your}, even revealing attributes like accent or emotional state—without identity—can constitute a serious privacy breach. Our experiments on Wav2Vec2, HuBERT, and Whisper across five attributes (gender, age, accent, emotion, and dysarthria) reveal substantial leakage of some attribute but not others, and our analysis provides  evidence that leakage is correlated with attribute coverage and representation in the base model.

\section{Methodology}
\label{section-methodology}
\vspace{-0.5em}

\subsection{Threat Model}
\label{sec:threat-model}
We consider a passive adversary at the server side of an FL system. The attacker does not interfere with training but performs a white-box attack, leveraging access to the global model \( W_g \) (distributed before personalization) and the locally trained model \( W_s \) for a target speaker \( s \), obtained after fine-tuning \( W_g \) on local speech. The attacker has no access to raw audio, transcripts, or metadata. We assume each model is fine-tuned on a single utterance per speaker, with personalization occurring locally. Given \( W_g \) and updated weights \( W_s \), the attacker's goal is to infer \textit{private and protected attributes} using only these weights. To support the attack, the adversary leverages public datasets to simulate fine-tuning on speech with known attributes. These simulations (shadow models~\cite{shokri2017membership}) provide labeled training data for auxiliary classifiers mapping weight updates to attributes. This reflects a realistic threat model in which the attacker follows the FL protocol but extracts sensitive information from updates.

\subsection{Attribute Inference Attack Model}
\label{sec:attack}
To test attribute inference feasibility, we simulate a passive attacker building a labeled dataset for the target attribute (e.g., gender) using public datasets. We collect labeled samples \( \{(x_i, y_i)\}_{i=1}^{n} \), with \( y_i \in \{\text{male}, \text{female}\} \), and fine-tune \( W_g \) on each sample \( x_i \), yielding \( n \) shadow models \( W_i, i=1,\dots,n \). For each \( W_i \), we extract summary statistics—mean, standard deviation, minimum, and maximum—from each parameter tensor \( p \in W_i \), concatenating them into a fixed-length feature vector \( z_i \in \mathbb{R}^d \):

\vspace{-0.5em}
\[
z_i = \text{concat}\left( \left[ \mu_p, \sigma_p, \min(p), \max(p) \right] \ \text{for each} \ p \in W_i \right)
\]
Here, \( d \) is the total dimensionality, determined by the number of parameter tensors.\footnote{For example, Whisper$_{small}$ (244M parameters) is reduced to $d=1916$.} We then compute class centroids (e.g., \( \bar{z}_m \) for male, \( \bar{z}_f \) for female) by averaging the feature vectors of shadow models within each class:

\vspace{-0.2em}
\[
\bar{z}_c = \frac{1}{N_c} \sum_{i=1}^{N_c} z_i \quad \text{where } z_i \in \text{class } c
\]

To infer the attribute of a new, unseen model from a target speaker, \( W_s \), the attacker extracts its feature vector \( z_s \) in the same way and compares it to the class centroids using normalized Euclidean distance. The predicted class corresponds to the closest centroid:
\vspace{-0.2em}
\begin{equation}
\hat{c} = \arg\min_{c} \left( \frac{ \| z_s - \bar{z}_c \|_2 }{ \| z_s \|_2 \cdot \| \bar{z}_c \|_2 } \right)
\end{equation}
This approach scales naturally to multi-class settings.

\section{Experiments}
The proposed attack is evaluated on three personal attributes: gender, age, and accent (correlated with country of origin)—as well as two speech-related conditions: emotional state and the presence of a speech disorder (dysarthria).

\subsection{Datasets}
We use publicly available speech datasets containing the target attributes. For \textit{gender}, \textit{age}, and \textit{accent}, we use the \textbf{Speech Accent Archive} (\textsc{SAA})~\cite{weinberger2011speech}, which provides controlled recordings of speakers reading the same scripted English sentence, ideal for isolating speaker traits. For \textit{speech disorders}, we use the \textbf{TORGO} dataset~\cite{rudzicz2012torgo}, with recordings from 15 speakers—8 with dysarthria (due to cerebral palsy or amyotrophic lateral sclerosis) and 7 without. For \textit{emotions}, we use the \textbf{RAVDESS} dataset~\cite{livingstone2018ryerson}, containing acted emotional speech from 24 professional speakers (12 female, 12 male) reading matched statements in a North American accent.

\subsection{Experimental Conditions}\label{sec:conditions}
We evaluate five attributes in total, each designed as a binary detection task. For \textbf{gender detection}, we use 200 native-English speakers from \textsc{SAA}, comprising 100 male and 100 female speakers, with a 75/25 train-test split, balanced by gender. The \textbf{age detection} task is also framed as a binary classification between two well-separated age ranges, 18–24 and 35–44 years, to avoid ambiguous boundary cases since age is continuous. We used 70 male native-English speakers (35 per class) from \textsc{SAA} and report results using five-fold cross-validation. For binary \textbf{Accent detection}, we used 200 speakers from \textsc{SAA}, equally divided between native and non-native English speakers, with a 75/25 split. \textbf{Emotion detection} consists of three binary tasks from \textsc{RAVDESS}—\textit{Calm} versus \textit{Angry}, \textit{Happy} versus \textit{Sad}, and \textit{Calm} versus \textit{Fearful}—each based on one fixed statement from 24 speakers (48 utterances) and evaluated with 6-fold cross-validation. For speech disorder detection, we select 10 utterances per speaker from \textsc{TORGO}, prioritizing longer ones. This is the only task with varying linguistic content, and it is evaluated using leave-one-speaker-out cross-validation.\footnote{Cross-validation reduces random effects but cannot eliminate systematic linguistic differences.}

\subsection{ASR Models}
We evaluate three widely used self-supervised speech models: \textbf{Wav2Vec2}-Base,\footnote{\url{huggingface.co/facebook/wav2vec2-base-960h}} which has 95 million parameters and was pre-trained and fine-tuned for ASR on 960 hours of LibriSpeech \cite{baevski2020wav2vec}; \textbf{HuBERT}-Large,\footnote{\url{huggingface.co/facebook/hubert-large-ls960-ft}} which has 300 million parameters and was also trained and fine-tuned on 960 hours of LibriSpeech \cite{hsu2021hubert}; and \textbf{Whisper}-Small,\footnote{\url{huggingface.co/openai/whisper-small}} which has 244 million parameters and was trained on 680,000 hours of multilingual and multitask labeled data \cite{radford2023robust}. LibriSpeech \cite{panayotov2015librispeech} is a corpus of read English speech by native adult speakers, both male and female, derived from audiobooks.
\vspace{-0.4em}

\section{Results}
\begin{figure}
    \centering
    \includegraphics[width=0.9\linewidth]{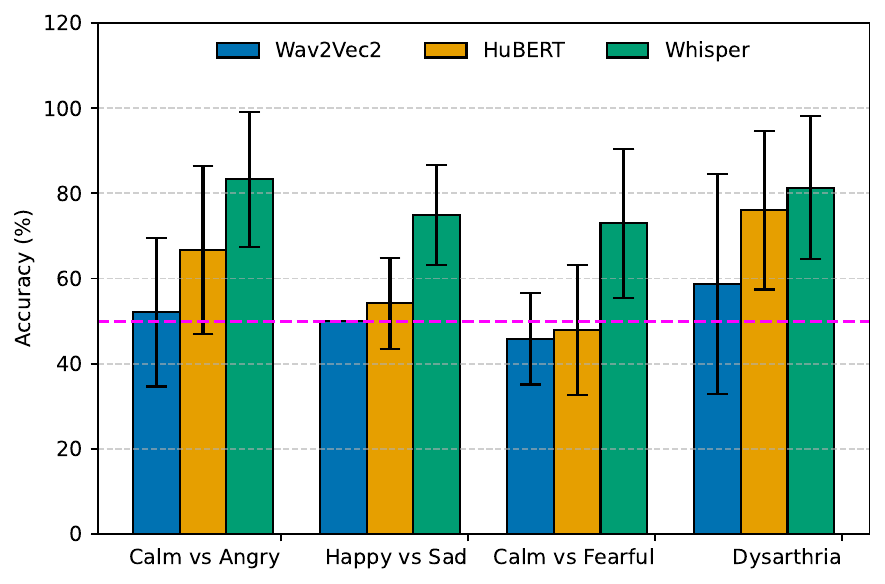}
    \caption{Mean accuracy for binary classification of Emotion \& Dysarthia with 95\% confidence intervals.}
    \label{fig:cv_results}
\end{figure}

We evaluated the attribute inference attack (Section~\ref{sec:attack}) for each binary condition (Section~\ref{sec:conditions}) using Wav2Vec2, HuBERT, and Whisper as base ASR models. Table~\ref{tab:model_comparison} reports the mean classification accuracy (attack success rate) with cross-validation applied where relevant. Gender was the most difficult attribute to predict, with accuracies ranging from 46\% to 64\%. Accent and age exhibited the strongest leakage, reaching 80–100\% accuracy across models; Wav2Vec2 achieved 100\% for both. Whisper consistently leaked attributes at $>$70\% accuracy, except for gender. Among emotions, anger yielded the highest detection accuracy, while other categories showed more variability. Variability in cross-validated results was assessed using standard error and 95\% confidence intervals (t-distribution), shown in Figure~\ref{fig:cv_results} (chance level: dashed magenta line). Age detection results were statistically significant for all models (CIs $\ge$ 95\%), so they are omitted from the figure. Whisper produced the most reliable results for all emotion categories and dysarthria despite small sample sizes, whereas Wav2Vec2 and HuBERT often performed near chance. Overall, these findings confirm that model updates can reveal sensitive attributes, with leakage strength depending on both the attribute and the model architecture.

\begin{table}
\centering
\renewcommand{\arraystretch}{1.1}
\resizebox{0.9\linewidth}{!}{%
\begin{tabular}{|l|c|c|c|}
\hline
\textbf{Task} & \textbf{Wav2Vec2} & \textbf{HuBERT} & \textbf{Whisper} \\
\hline
\textbf{Gender:} Male / Female & 64\% & 63\% & 46\% \\
\textbf{Age:} 18-24 / 35-44 & 100\% & 97\% & 94\% \\
\textbf{Accent:} Native / Accented & 100\% & 80\% & 93\% \\
\textbf{Dysarthria:} True / False & 59\% & 76\% & 81\% \\
\textbf{Emotion}: Calm / Angry & 52\% & 67\% & 83\% \\
\textbf{Emotion:} Happy / Sad & 50\% & 54\% & 75\% \\
\textbf{Emotion:} Calm / Fearful & 46\% & 48\% & 73\% \\
\hline
\end{tabular}
}
\caption{Attack success rate in terms of accuracy (\%)}
\label{tab:model_comparison}
\end{table}

\section{Per-Layer Analysis}
\begin{figure}
    \centering
    \includegraphics[width=\linewidth]{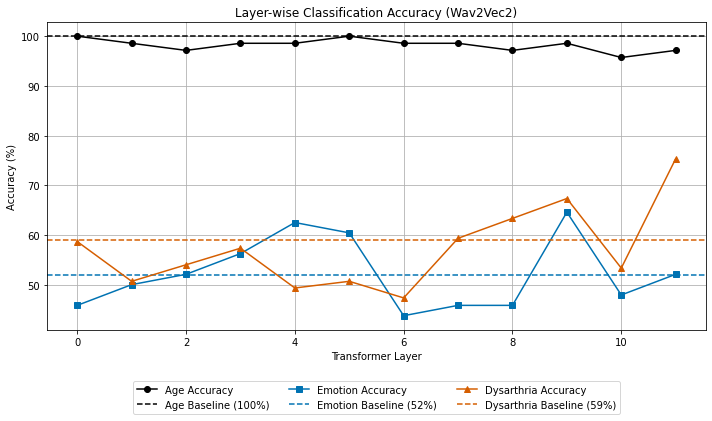}
    \caption{Layer-wise classification accuracy using Wav2Vec2 encoder layers for three tasks. The dashed lines show the result obtained using the full set of parameters.}
    \label{fig:layerwise}
\end{figure}

We analyzed where attribute-specific information is encoded in Wav2Vec2 by applying the attribute inference attack (Section~\ref{sec:attack}) to parameter tensors from individual transformer layers. Accuracy was evaluated for age, emotion (calm vs. angry), and dysarthria, with results in Figure~\ref{fig:layerwise} (mean accuracy across cross-validation folds). Age detection remained consistently high across all layers, indicating that effective attacks may be implemented using only a subset of parameters. In contrast, emotion and dysarthria showed greater variability, with certain mid-to-late layers 
outperforming full-model inference. These findings suggest that attribute-specific information is distributed unevenly across layers, and that targeting specific layers can sometimes enhance inference accuracy.

\section{Fine-Grained Accent Classification}
\label{sec:investigating-accent-leakage}
\begin{figure}[h]
    \centering
    \begin{subfigure}[b]{0.4\textwidth}
        \centering
        \caption{Before (incorrect predictions in magenta).}
        \label{fig:multi_accent_large}\hspace{-3em}\includegraphics[width=\textwidth]{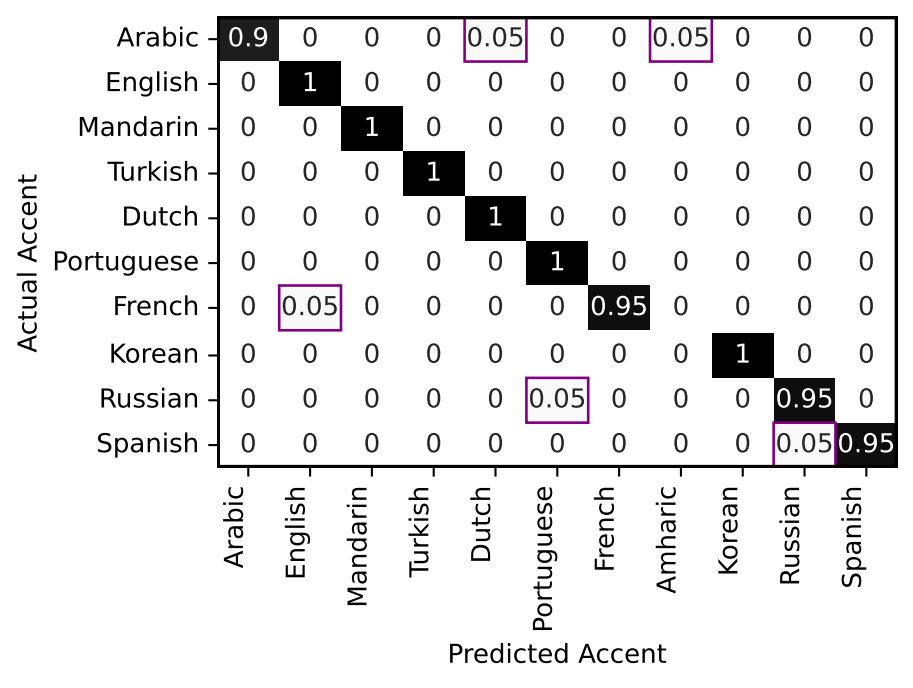}
        
    \end{subfigure}
    \hfill
    \begin{subfigure}[b]{0.4\textwidth}
        \centering
        \caption{After (correct predictions in green).}
        \label{fig:defense-FT-results}
        \hspace{-3em}
        \includegraphics[width=\textwidth]{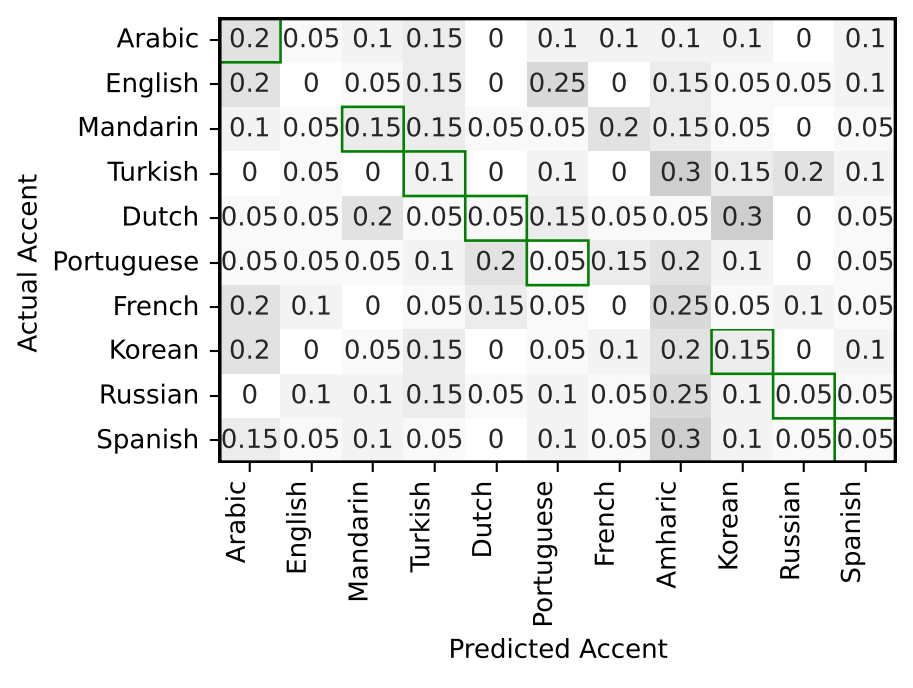}
        
    \end{subfigure}
    \caption{Wav2Vec2 accent classification results before and after fine-tuning. Results are shown as proportions of the total for each class, so the diagonal corresponds to precision. Note: Amharic is only used in training the attribute classifier, with no test samples. }
    \label{fig:Wav2Vec 2 Multi-Class Accent Prediction Results}
\end{figure}

We further analyze accent prediction, as it yielded some of the most consistent binary classification results. The \textsc{SAA} corpus provides controlled recordings from speakers with diverse native accents, enabling multi-class classification. We use Wav2Vec2, which achieved perfect binary accuracy, and select the ten most represented accents—English, Spanish, Arabic, Dutch, Mandarin, French, Korean, Portuguese, Russian, and Turkish—plus Amharic (training only). For each accent, 15 samples are used for training and 20 unseen speakers for testing; Amharic has no test set due to limited data. As shown in Figure~\ref{fig:multi_accent_large}, the attack achieves $\ge 90\%$ accuracy for all tested accents\footnote{Similar results were observed for HuBERT and Whisper.}.  

\subsection{Effect of Targeted Fine-Tuning}
One observation to note from the results so far is the high disparity in detection accuracy between different attributes. Gender, for instance, was predicted at near chance levels using FL model weights, in spite of it being one of the most distinguishable features in raw speech \cite{leung2018voice}. On the other hand, accent was predicted almost perfectly, even in multi-class setting. A likely factor behind the high leakage is the mismatch between ASR pre-training data (e.g., LibriSpeech) and target-speaker accents, which can induce larger and more distinct gradients due to higher surprisal. Another possible cause is catastrophic forgetting \cite{french1999catastrophic}, where attributes included in early training but unseen in later updates still produce high surprisals. To test this, we fine-tune the global model on previously unused speakers of the ten most frequent accents in \textsc{SAA} (up to 20 sample per accent, total 137 samples). After fine-tuning, the attack’s success drops sharply, with no accent exceeding 0.2 accuracy (Figure~\ref{fig:defense-FT-results}). This confirms that limited accent diversity in training amplifies leakage, and that training on a wide range of accents not only serves inclusion \cite{maison2023improving} but also mitigates accent-based attribute inference. This may also explain the negative results for gender, as gender variations are well represented in the underlying models. 

\subsection{Generalization to Unseen Accents}
Fine-tuning on accented speech can defend against our attack for known accents, but what about unseen accents? We test this by selecting five accents from \textsc{SAA} absent in prior experiments: Japanese, Italian, German, Polish, and Macedonian. For each, we use six additional samples for training shadow models and 14 unseen speakers for testing. As shown in Table~\ref{tab:accent_generalization_unseen}, the attack achieves high precision and recall across all accents. These results indicate that accents not adequately represented in ASR pre-training or fine-tuning data remain highly vulnerable, as their acoustic characteristics yield more distinguishable parameter updates. Partial coverage may therefore be insufficient for privacy preservation, particularly during initial client enrollment when novel attributes are more likely to appear.

\begin{table}
\centering
\renewcommand{\arraystretch}{1.1}
\small 
\scalebox{0.9}{
\begin{tabular}{|l|c|c|c|}
\hline
\textbf{Accent} & \textbf{Precision} & \textbf{Recall} & \textbf{F1-score} \\
\hline
Japanese & 1.00 & 1.00 & 1.00 \\
Italian & 0.81 & 0.93 & 0.87 \\
German & 0.85 & 0.79 & 0.82 \\
Polish & 1.00 & 1.00 & 1.00 \\
Macedonian & 1.00 & 1.00 & 1.00 \\
\hline
\end{tabular}
}
\caption{Wav2Vec2 Accent Classification - Generalization to Unseen Accents}
\label{tab:accent_generalization_unseen}
\end{table}
\vspace{-1em}

\section{Conclusion}
This study presents an investigation into attribute leakage in federated Automatic Speech Recognition (ASR). We show attack success rates for age, accent, dysarthria, and emotion detection that exceed chance levels, particularly using the Whisper model as the base ASR system. Age and accent, in particular, are highly vulnerable to such attribute attacks. Inferring such attributes–particularly in combination–can enable profiling, discrimination, and surveillance. For example, a speech disorder label combined with IP/geolocation data can uniquely identify a user. While it is understood that FL improves privacy by keeping audio on-device, our work highlights that this protection is uneven—certain traits (e.g., accent and age) can be inferred successfully more than others. We also show experimentally that this leakage may be explained by the lack of coverage of these traits in the training data of the underlying models. When models were exposed to even a small, diverse sample of accented speech prior to client updates, the effectiveness of the leakage attack significantly declined for the covered accents. This suggests that pre-training on diverse demographies and conditions prior to federated learning could result in better robustness against privacy attacks in FL settings. Finally, our controlled experiments demonstrate privacy leakage using minimal attribute training data, but results on gender detection also show that complex and interdependent attributes may be hard to detect in real-world scenarios, especially if these attributes are well represented in the model's initial training data prior to FL setup. This work reveals previously undocumented vulnerabilities in federated ASR models and demonstrates a potential approach to improve their security.

\bibliographystyle{IEEEbib}
\bibliography{main}

\end{document}